\documentclass[10pt,twocolumn]{article} 
\usepackage{iccv}
\usepackage{times}
\usepackage{graphicx}
\usepackage{amsmath,amssymb,amsfonts,amsthm}
\usepackage{url,hyperref}
\usepackage{array}
\usepackage{color}
\usepackage{enumitem}
\usepackage{float}
\usepackage{mathtools}
\usepackage{setspace}

\theoremstyle{definition}
\newtheorem{definition}{Definition}[section]

\newcolumntype{L}[1]{>{\raggedright\let\newline\\\arraybackslash\hspace{0pt}}m{#1}}
\newcolumntype{C}[1]{>{\centering\let\newline\\\arraybackslash\hspace{0pt}}m{#1}}
\newcolumntype{R}[1]{>{\raggedleft\let\newline\\\arraybackslash\hspace{0pt}}m{#1}}

\setlist[itemize]{noitemsep, topsep=0pt}
\setlist[enumerate]{noitemsep, topsep=0pt}

\iccvfinalcopy

\begin{document}

\title{STEP: Spatial Temporal Graph Convolutional Networks for Emotion Perception from Gaits}

\author{Uttaran Bhattacharya$^1$, Trisha Mittal$^1$, Rohan Chandra$^1$, Tanmay Randhavane$^2$, \\
Aniket Bera$^1$, Dinesh Manocha$^1$ \\
$^1$ Department of Computer Science, University of Maryland, College Park, USA \\
$^2$ Department of Computer Science, University of North Carolina, Chapel Hill, USA \\
\url{https://gamma.umd.edu/researchdirections/affectivecomputing/step}  \\
}

\maketitle
\thispagestyle{empty}

\begin{abstract}
We present a novel classifier network called STEP, to classify perceived human emotion from gaits, based on a Spatial Temporal Graph Convolutional Network (ST-GCN) architecture. Given an RGB video of an individual walking, our formulation implicitly exploits the gait features to classify the emotional state of the human into one of four emotions: happy, sad, angry, or neutral. We use hundreds of annotated real-world gait videos and augment them with thousands of annotated synthetic gaits generated using a novel generative network called STEP-Gen, built on an ST-GCN based Conditional Variational Autoencoder (CVAE). We incorporate a novel push-pull regularization loss in the  CVAE formulation of STEP-Gen to generate realistic gaits and improve the classification accuracy of STEP.
We also release a novel dataset (E-Gait), which consists of $2,177$ human gaits annotated with perceived emotions along with thousands of synthetic gaits. In practice, STEP can learn the affective features and exhibits classification accuracy of $89\%$ on E-Gait, which is $14 - 30\%$  more accurate over prior methods.
\end{abstract}

\section{Introduction}\label{sec:intro}
Human emotion recognition using intelligent systems is an important socio-behavioral task that arises in various applications, including behavior prediction~\cite{behavior2}, surveillance~\cite{surveillance2}, robotics~\cite{bauer2009autonomous}, affective computing~\cite{affective1,affective2}, etc. Current research in perceiving human emotion predominantly uses facial cues~\cite{facial_deep2}, speech~\cite{prosodic2}, or physiological signals such as heartbeats and respiration rates~\cite{physiological2}. These techniques have been used to identify and classify broad emotions including happiness, sadness, anger, disgust, fear and other combinations~\cite{ekman1967head}.

Understanding the perceived emotions of individuals using non-verbal cues, such as face expressions or body movement, is regarded as an important and challenging problem in both AI and psychology, especially when self-reported emotions are unreliable or misleading~\cite{perception_imp}. Most prior work has focused on facial expressions, due to the availability of large datasets~\cite{AU}. However, facial emotions can be unreliable in contexts such as referential expressions~\cite{face_re} or the presence or absence of an audience~\cite{face_audience}. Therefore, we need better techniques that can utilize other non-verbal cues.

\begin{figure}[t]
    \centering
    \includegraphics[width=\columnwidth]{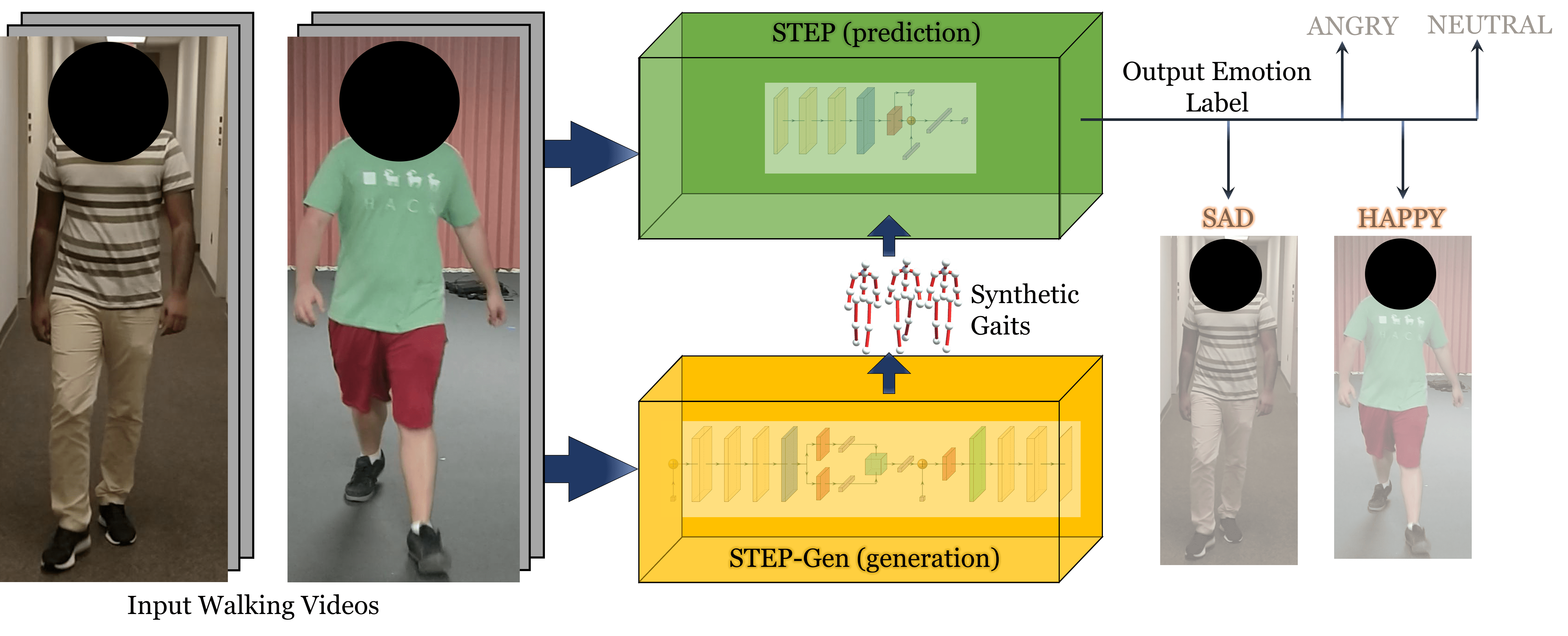}
    \caption{\small\textbf{STEP and STEP-Gen:} We present a novel classifier network (STEP) to predict perceived emotions from gaits, as shown for this walking video.  Furthermore, we present a generator network (STEP-Gen) to generate annotated synthetic gaits from our real world gait dataset to improve the accuracy of STEP.
    We evaluate their performance on a novel E-Gait dataset and observe $14-30\%$ improvement in the classification accuracy over prior methods. }
    \label{fig:cover}
\end{figure}

In this paper, we mainly focus on using movement features corresponding to gaits in a walking video for emotion perception.  A gait is defined as an ordered temporal sequence of body joint transformations (predominantly translations and rotations) during the course of a single walk cycle. Simply stated, a person's gait is the way the person walks. Prior work in psychology literature has reported that participants were able to identify sadness, anger, happiness, and pride by observing affective features corresponding to arm swinging, long strides, erect posture, collapsed upper body, etc.~\cite{gait_psych1,gait_psych2,gait_psych3,gait_psych4}.

There is considerable recent work on pose or gait extraction from a walking video using deep convolutional network architectures and intricately designed loss functions~\cite{pose_extract_1,pose_extract_2}.
Gaits have also been used for a variety of applications including action recognition~\cite{acrec2,stgcn} and person identification~\cite{identification2}. However, the use of gaits for automatic emotion perceptions has been fairly limited, primarily due to a lack of gait data or videos annotated with emotions~\cite{gait_emo}. It is difficult and challenging to generate a large dataset with many thousands of annotated  real-world gait videos to train a network.

\noindent{\bf Main Results:} We present a learning-based approach to classify perceived emotions of an individual walking in a video. Our formulation consists of a novel classifier and a generative network as well as an annotated gait video dataset. The main contributions include:
\begin{enumerate}
\item A novel end-to-end Spatial Temporal Graph Convolution-Based Network (STEP), which implicitly extracts a person's gait from a walking video to predict their emotion. STEP combines deeply learned features with affective features to form hybrid features.

\item A Conditional Variational Autoencoder (CVAE) called STEP-Gen, which is trained on a sparse real-world annotated gait set and can easily generate  thousands of annotated synthetic gaits. We enforce the temporal constraints~(\textit{e.g.,} gait drift and gait collapse) inherent in gaits directly into the loss function of the CVAE, along with a novel push-pull regularization loss term. Our formulation helps to avoid over-fitting by generating more realistic gaits.  These synthetic gaits improve the accuracy of STEP by $6\%$ in our benchmarks.

\item We present a new dataset of human gaits annotated with emotion labels (E-Gait). It consists of $2,177$ real-world and $4,000$ synthetic gait videos annotated with different emotion labels.

\end{enumerate}
We have evaluated the performance of STEP on E-Gait. The gaits in this dataset were extracted from videos of humans walking in both indoor and outdoor settings and labeled with one of four emotions: angry, sad, happy, or neutral. In practice, STEP results in classification accuracy of $87\%$ on E-Gait. We have compared it with prior methods and observe:
\begin{itemize} 
\item An accuracy increase of $14\%$ over prior learning-based method~\cite{tanmay_lstm}. This method uses LSTMs for modeling their input, but for an action recognition task.
\item Accuracy improvement of $21-30\%$ on the absolute over prior gait-based emotion recognition methods reported in the psychology literature that use affective features.
\end{itemize}

\section{Related Work}\label{sec:rw}
We provide a brief overview of prior work in emotion perception and generative models for gait-like datasets.

\noindent\textbf{Emotion Perception.} Face and speech data have been widely used to perceive human emotions. Prior methods that use faces as input commonly track action units on the face such as points on the eyebrow, cheeks and lips~\cite{AU}, or track eye movements~\cite{eye2} and facial expressions~\cite{facial}. Speech-based emotion perception methods use either spectral features or prosodic features like loudness of voice, difference in tones and changes in pitch~\cite{prosodic2}. With the rising popularity of deep learning, there is considerable work on developing learned features for emotion detection from large-scale databases of faces~\cite{facial_deep3,facial_deep4} and speech signals~\cite{speech_deep3}. Recent methods have also looked at the cross-modality of combined face and speech data to perform emotion recognition~\cite{crossmodal2}. In addition to faces and speech, physiological signals such as heartbeats and respiration rates~\cite{physiological2} have also been used to increase the accuracy of emotion perception. Our approach for emotion perception from walking videos and gaits is complimentary to these methods and can be combined.

Different methods have also been proposed to perceive emotions from gaits. Karg et al.~\cite{karg2010recognition} use PCA-based classifiers, and Crenn et al.~\cite{crenn2016body} use SVMs on affective features. Venture et al.~\cite{venture2014recognizing} use autocorrelation matrices between joint angles to perform similarity-based classification. Daoudi et al.~\cite{daoudi2017emotion} represent joint movements as symmetric positive definite matrices and perform nearest neighbor classification. 

Gaits have also been widely used in the related problem of action recognition~\cite{c3d,i3c,twostream,tcn,stgcn}. In our approach, we take motivation from prior works on both, emotion perception and action recognition from gaits.

\noindent\textbf{Gait Generation.} Collecting and compiling a large dataset of annotated gait videos is indeed a challenging task. As a result, it is important to develop generative algorithms for gaits conditioned on emotion labels. Current learning-based generation models are primarily based on Generative Adverserial Networks~(GANs) or Variational Autoencoders~(VAEs). MoCoGAN~\cite{mocogan} uses a GAN-based model, the latent space of which is divided into motion space (for generating temporal features) and content space (for generating spatial features). It can generate tiny videos of facial expressions corresponding to various emotions. vid2vid~\cite{vid2vid} is a state-of-the-art GAN-based network that uses a combined spatial temporal adversarial objective to generate high-resolution videos, including videos of human poses and gaits when trained on relevant real data. Other generative methods for gaits learn the initial poses and the intermediate transformations between frames in separate networks, and then combine the generated samples from both networks to develop realistic gaits~\cite{pose_guided_gait1,pose_guided_gait2}. In this work, we model gaits as skeletal graphs and use spatial-temporal graph convolutions~\cite{stgcn} inside a VAE to generate synthetic gaits.

\section{Background}\label{sec:gait}
In this section, we give a brief overview of Spatial Temporal Graph Convolutional Networks (ST-GCNs) and Conditional Variational Autoencoders (CVAE). 
\subsection{GCN and ST-GCN}
\label{subsec:stgcn}
The Graph Convolutional Network (GCN) was first introduced in~\cite{gcn1} to apply convolutional filters to arbitrarily structured graph data. Consider a graph $\mathcal{G}$=$\{\mathcal{V}, \mathcal{E}\}$ with $N$ = $\lvert\mathcal{V}\rvert$ nodes. Also consider a feature matrix $X \in \mathbb{R}^{N \times F}$, where row $x_i \in \mathbb{R}^F$ corresponds to a feature for vertex $i$. The propagation rule of a GCN is given as
\begin{equation}
     Z^{(l+1)} = \sigma(AZ^{(l)}W^{(l)}),
\end{equation}

where $Z^{(l)}$ and $Z^{(l+1)}$ are the inputs to the $l$-th and the $(l+1)$-th layers of the network, respectively.  $Z^{(0)}$=$X$, $W^{(l)}$ is the weight matrix between the $l$-th and the $(l+1)$-th layers, $A$ is the $N\times N$ adjacency matrix associated with the graph $\mathcal{G}$ and $\sigma(\cdot)$ is a non-linear activation function (\textit{e.g.}, ReLU). Thus, a GCN takes in a feature matrix $X$ as an input and generates another feature matrix $Z^{(L)}$ as the output, $L$ being the number of layers in the network. In practice, each weight matrix $W$ in a GCN represents a convolutional kernel. Multiple such kernels can be applied to the input of a particular layer to get a feature tensor as output, similar to a conventional Convolutional Neural Network (CNN). For example, if $K$ kernels, each of dimension $F\times D$ are applied to the input $X$, then the output of the first layer will be an $N\times D\times K$ feature tensor.

Yan et al.~\cite{stgcn} extended GCNs to develop the spatial temporal GCN (ST-GCN), which can be used for action recognition from human skeletal graphs. The graph in their case is the skeletal model of a human extracted from videos. Since they extract poses from each frame of a video, their input is a temporal sequence of such skeletal models. ``Spatial'' refers to the spatial edges in the skeletal model, which are the limbs connecting the body joints. ``Temporal'' refers to temporal edges which connect the positions of each joint across different time steps. Such a representation enables the gait video to be expressed as a single graph with a fixed adjacency matrix, and thus can be passed through a GCN network. The feature per vertex in their case is the 3D position of the joint represented by that vertex. In our work, we use the same representation for gaits, described later in Section~\ref{subsec:gaits_from_videos}.

\subsection{Conditional Variational Autoencoder}
\label{subsec:cvae}
The variational autoencoder~\cite{vae} is an encoder-decoder architecture that is used for data generation based on Bayesian inference. The encoder transforms the training data into a latent lower-dimensional distribution space. The decoder draws random samples from that distribution and generates synthetic data that are as similar to the training data as possible.

In conditional VAE~\cite{cvae}, instead of generating from a single distribution space learned by the encoder, it learns separate distributions for the separate classes in the training data. Thus, given a class, the decoder produces random samples from the conditional distribution of that class, and generates synthetic data of that class from those samples. Furthermore, if we assume that the decoder generates Gaussian variables for every class, then the negative log likelihood for each class is given by the MSE loss
\begin{equation}
     \mathcal{L}_o = \lVert x - f_{\theta_c}(z) \rVert^2
\end{equation}
where $f_{\theta_c}(\cdot)$ denotes the decoder function for class $c$, $x$ represents the training data, and $z$ the latent random variable. We incorporate a novel push-pull regularization loss on top of this standard CVAE loss, as described in Section~\ref{subsec:metric}.

\begin{figure*}[t]
    \centering
    \includegraphics[width=\linewidth]{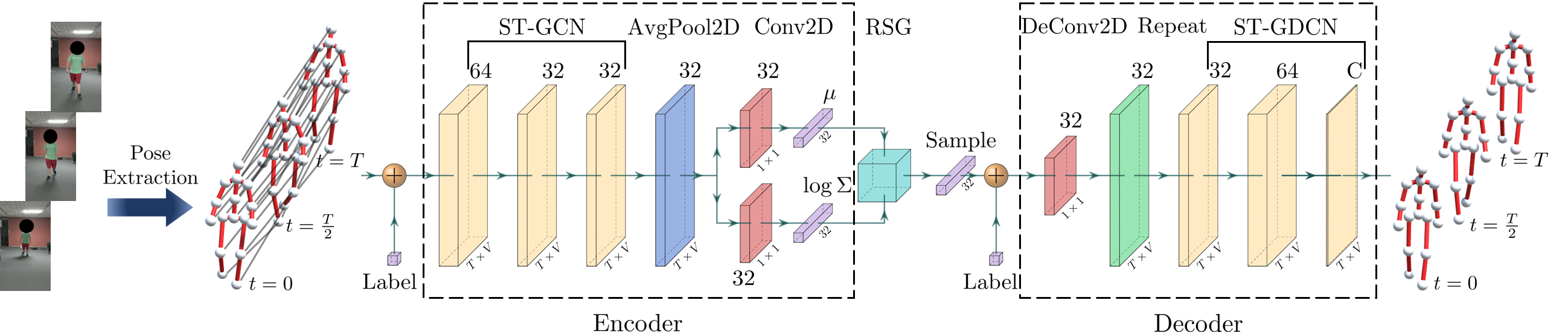}
    \caption{\small{\bf Our Generation Network (STEP-Gen):} The encoder consists of ST-GCN, Average Pool and Conv2D layers. The decoder consists of DeConv2D, Repeat and ST-GDCN layers. RSG (Random Sample Generator) is used to generate random samples from the latent space. $+$ denotes appending; $T$: number of time steps ($75$ in our dataset); $V$: number of nodes ($16$ in our dataset); $C$: dimension of each node ($3$ in our dataset). Input: Human gaits processed from walking videos and corresponding emotion label. Spheres are nodes, thick red lines are spatial edges and thin gray lines are temporal edges. Output: Human gaits corresponding to the input label, with same $T$, $V$, and $C$.}
    \label{fig:gen_network}
\end{figure*}

\section{STEP and STEP-Gen}
Our objective is to perform emotion perception from gaits. Based on prior work~\cite{gait_psych4,karg2010recognition,crenn2016body}, we assume that emotional cues are largely determined by localized variances in gaits, such as swinging speed of the arm (movement of 3 adjacent joints: shoulder, elbow and hand), stride length and speed (movement of 3 adjacent joints: hip, knee and foot), relative position of the spine joint w.r.t. the adjacent root and neck joints and so on. Convolutional kernels are known to capture such local variances and encode them into meaningful feature representations for learning-based algorithms~\cite{cnn_alexnet}. Additionally, since we treat gaits as a periodic motion that consists of a sequence of localized joint movements in 3D, we therefore use GCNs for our generation and classification networks to capture these local variances efficiently. In particular, we use Spatial Temporal GCNs (ST-GCNs) developed by~\cite{stgcn} to build both our generation and classification networks. We now elaborate our entire approach in detail.

\subsection{Extracting Gaits from Videos}
\label{subsec:gaits_from_videos}
Naturally collected human gait videos contain a wide variety of extraneous information such as attire, items carried (\textit{e.g.}, bags or cases), background clutter, etc. We use a state-of-the art pose estimation method~\cite{pose_extract_1} to extract clean, 3D skeletal representations of the gaits from videos. 
Moreover, gaits in our dataset are collected from varying viewpoints and scales. To ensure that the generative network does not end up generating an extrinsic mean of the input gaits, we perform view normalization. Specifically, we transform all gaits to a common point of view in the world coordinates using the Umeyama method~\cite{umeyama}.
Thus, a gait in our case is a temporal sequence of view normalized skeletal graphs extracted per frame from a video. We now provide a formal definition for gait. 
\begin{definition}
A gait is represented as a graph $\mathcal{G} = (\mathcal{V}, \mathcal{E})$, where $\mathcal{V}$ denotes the set of vertices and $\mathcal{E}$ denotes the set of edges, such that
\begin{itemize}
    \item $v_i^t \in \mathcal{V}$, $i \in \{1, \dots V\}$ represents the 3D position of the $i$-th joint in the skeleton at time step $t$ and $V$ is the total number of joints in the skeleton.
    \item $\mathcal{A}_i^t \subseteq \mathcal{V}$ is the set of all nodes that are adjacent to $v_i^t$ as per the skeletal graph at time step $t$,
    \item $v_i \coloneqq \{v_i^t\}_{t \in \{1, \dots, T\}}$ denotes the set of positions of of the $i$-th joint across all time steps $1 \dots T$,
    \item $(v_i^t, v_j^t) \in \mathcal{E}$, $\forall v_j^t \in \mathcal{A}_i^t \cup v_i$, $\forall t \in \{1, \dots, T\}$, $\forall i \in \{1, \dots, V\}$.
\end{itemize}
\label{def:gait}
\end{definition}
\indent A key pre-requisite for using GCNs is to define the adjacency between the nodes in the graph~\cite{gcn1,gcn2,stgcn}. Note that as per definition~\ref{def:gait}, given fixed $T$ and $V$, any pair of gaits $\mathcal{G}_x$ and $\mathcal{G}_y$ can have different sets of vertices, $\mathcal{V}_x$ and $\mathcal{V}_y$ respectively, but necessarily have the same edge set $\mathcal{E}$ and hence the same adjacency matrix $A$. This useful property of the definition allows us to maintain a unique notion of adjacency for all the gaits in a dataset, and thus develop ST-GCN-based networks for the dataset.

\subsection{STEP-Gen: The Generation Network}
\label{subsec:step-gen}
We show our generative network in Figure~\ref{fig:gen_network}. Our network architecture is based on the Conditional Variational Autoencoder (CVAE)~\cite{cvae}.

In the encoder, each $C\times T\times V$ dimensional input gait, pre-processed from a video (as per Section~\ref{subsec:gaits_from_videos}), is appended with the corresponding label, and passed through a set of 3 ST-GCN layers~(yellow boxes). $C$=$3$ is the feature dimension of each node in the gait, representing the 3D position of the corresponding joint. The first ST-GCN layer has $64$ kernels and the next two have $32$ kernels each. The output from the last ST-GCN layer is average pooled along both the temporal and joint dimensions~(blue box). Thus, the output of the pooling layer is a $32\times 1 \times 1$ tensor. This tensor is passed through two $1\times 1$ convolutional layers in parallel~(red boxes). The outputs of the two convolutional layers are $32$ dimensional vectors, which are the mean and the log-variance of the latent space respectively~(purple boxes). All ST-GCN layers are followed by the ReLU nonlinearity, and all the layers are followed by a BatchNorm layer (not shown separately in Figure~\ref{fig:gen_network}).

In the decoder, we generate random samples from the $32$ dimensional latent space and append them with the same label provided with the input. As commonly performed in VAEs, we use the reparametrization trick~\cite{vae} to make the overall network differentiable. The random sample is passed through a $1\times 1$ deconvolutional layer~(red box), and the output feature is repeated (``un-pooled'') along both the temporal and the joint dimension~(green box) to produce a $32\times T\times V$ dimensional tensor. This tensor is then passed through 3 spatial temporal graph deconvolutional layers (ST-GDCNs)~(yellow boxes). The first ST-GDCN layer has $32$ kernels, the second one has $64$ channels, and the last one has $C$=$3$ channels. Hence, we finally get a $C\times T\times V$ dimensional tensor at the output, which is a synthetic gait for the provided label. As in the encoder part, all ST-GDCN layers are followed by a ReLU nonlinearity, and all layers are followed by a BatchNorm layer (not shown separately in Figure~\ref{fig:gen_network}).

Once the network is trained, we can generate new synthetic gaits by drawing random samples from the $32$ dimensional latent distribution space parametrized by the learned $\mu$ and $\Sigma$.

The original CVAE loss $\mathcal{L}_o$ is given by:
\begin{equation}
     \mathcal{L}_o = \sum_{t=1}^T \lVert v_R^t - v_S^t \rVert^2,
    \label{equation:originalCVAE}
\end{equation}
\noindent where $ v^t = \begin{bmatrix} v_1^t & \dots &  v_V^t\end{bmatrix}^\top$, where each $v_i^t$ is assumed to be a row vector consisting of the 3D position of the joint $i$ at frame $t$. The subscripts $R$ and $S$ stand for real and synthetic data respectively.

Each gait corresponds to a temporal sequence. Therefore, for any gait representation, it is essential to incorporate such temporal information. This is even more important as temporal changes in a gait provide significant cues for emotion perception~\cite{gait_psych4,karg2010recognition,crenn2016body}. But, the baseline-CVAE architecture does not take into account the temporal nature of the gaits. We therefore modify the original reconstruction loss of the CVAE by adding regularization terms that enforce the desired temporal constraints (Equation~\ref{equation:STEP-Gen Loss}).

We propose a novel ``push-pull'' regularization scheme. We first make sure that sufficient movement occurs in a generated gait across the frames so that the joint configurations at different time frames do not collapse into a single configuration. This is the ``push'' scheme. Simultaneously, we make sure that the generated gaits do not drift too far from the real gaits over time due to excessive movement. This is the ``pull'' scheme. 
\begin{itemize}
    \item \textit{Push}: We require the synthetic data to resemble the joint velocities and accelerations of the real data as closely as possible. The velocity $vel_i^t$ of a node $i$ at a frame $t$ can be approximated as the difference between the positions of the node at frames $t$ and $t-1$, \textit{i.e.},
    \begin{equation}
         vel_i^t = v_i^t - v_i^{t-1}
    \end{equation}
    Similarly, acceleration $acc_i^t$ of a node $i$ at a frame $t$ can be approximated as the difference between the velocities of the node at frame $t$ and $t-1$, \textit{i.e.},
    \begin{equation}
         acc_i^t = vel_i^t - vel_i^{t-1} = v_i^t - 2v_i^{t-1} + v_i^{t-2}
    \end{equation}
    We use the following loss for gait collapse:
    \begin{equation}
         \mathcal{L}_c = \sum_{t=2}^T \lVert vel_R^t-vel_S^t \rVert^2 + \sum_{t=3}^T \lVert acc_R^t-acc_S^t \rVert^2
        \textbf{}
    \end{equation}
    where $ vel^t =  \begin{bmatrix}  vel_1^t & \dots &  vel_V^t\end{bmatrix}^\top$ and $ acc^t =  \begin{bmatrix}  acc_1^t & \dots &  acc_V^t\end{bmatrix}^\top$.
    
    \item \textit{Pull}: When the synthetic gait nodes are enforced to have non-zero velocity and acceleration between the frames, the difference between the synthetic node positions and the corresponding real node positions tends to increase as the number of frames increases. This is commonly known as the drift error. In order to constrain this error, we use the notion of anchor frames. At the anchor frames, we impose additional penalty on the loss between the real and synthetic gaits. In order to be effective, we need to ensure that there are a high number of anchor frames and they are as far apart as possible. Based on this trade off, we choose 3 anchor frames in the temporal sequence --- the first frame, the middle frame and the last frame of the gait. We use the following loss function for gait drift:
    \begin{equation}
         \mathcal{L}_d = \sum_{t=1}^T\sum_{\omega \in \Omega} \lVert v_R^t-v_R^\omega - (v_S^t-v_S^\omega)\rVert^2
    \end{equation}
    where $\Omega$ denotes the set of anchor frames.
\end{itemize}
Finally, our modified reconstruction loss $\mathcal{L}_r$ of the CVAE is given by
\begin{equation}
     \mathcal{L}_r = \mathcal{L}_o + \lambda_c\mathcal{L}_c + \lambda_d\mathcal{L}_d
    \label{equation:STEP-Gen Loss}
\end{equation}
where $\lambda_c$ and $\lambda_d$ are the regularization weights. Note that this modified loss function still satisfies the ELBO bound~\cite{vae}, if we assume that the decoder generates variables from a mixture of Gaussian distributions for every class, with the original loss, the push loss ad the pull loss representing the 3 Gaussian distributions in the mixture.

\subsection{STEP: The Classification Network}
\label{subsec:classify}
\begin{figure}[t]
    \centering
    \includegraphics[width=\linewidth]{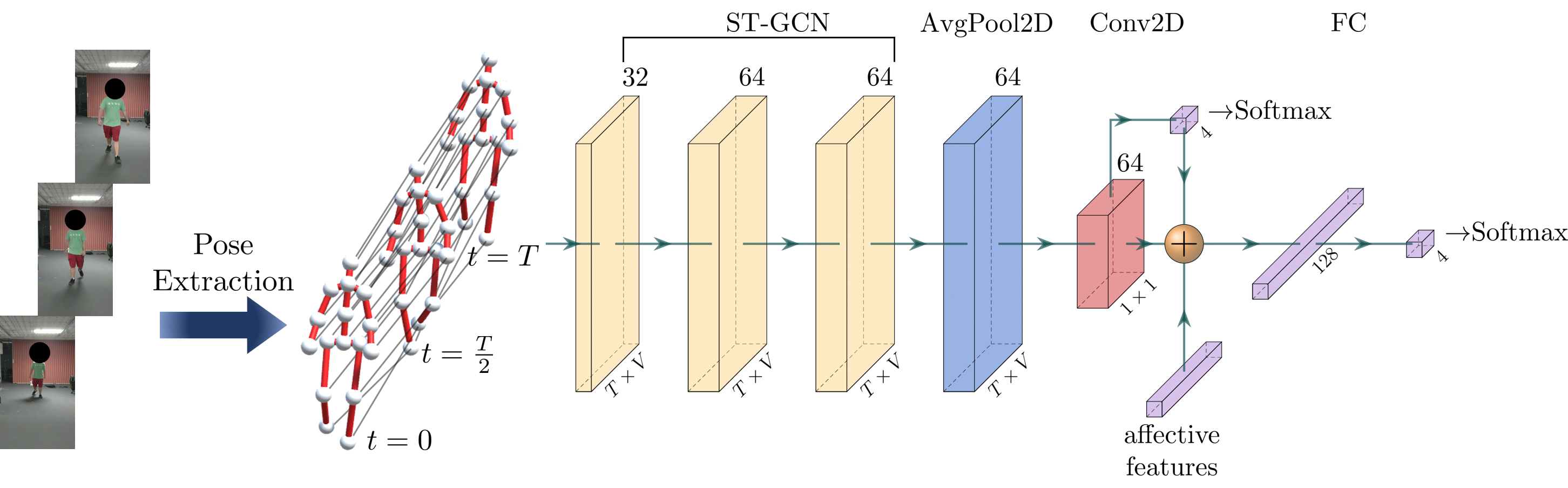}
    \caption{\small{\bf Our Classifier Network (STEP):} It consists of ST-GCN, Average Pool, Conv2D and fully connected (FC) layers. $+$ denotes appending. $T$: number of time steps ($75$ in our dataset); $V$: number of nodes ($16$ in our dataset); $C$: dimension of each node ($3$ in our dataset). Input: Human gaits processed from walking videos. Spheres are nodes, thick red lines are spatial edges and thin gray lines are temporal edges. Output: Predicted label after Softmax. The first Softmax from the left gives the output of Baseline-SETP, and the second Softmax gives the output of STEP.}
    \label{fig:classify_network}
\end{figure}
We show out classifier network in Figure~\ref{fig:classify_network}. In the base network, each input gait is passed through a set of $3$ ST-GCN layers~(yellow boxes). The first ST-GCN layer has $32$ kernels and the next two have $64$ kernels each. The output from the last ST-GCN layer is average pooled~(blue box) in both the temporal and joint dimensions and passed through a $1\times 1$ convolutional layer~(red box). The output of the convolutional layer is passed through a fully connected layer of dimension $4$ (corresponding to the $4$ emotion labels that we have), followed by a softmax operation to generate the class labels. All the ST-GCN layers are followed by the ReLU nonlinearity and all layers except the fully connected layer are followed by a BatchNorm layer (not shown separately in Figure~\ref{fig:classify_network}). We refer to this version of the network as the {\em Baseline-STEP}.

Prior work in gait analysis has shown that affective features for gaits provide important information for emotion perception~\cite{gait_psych4,karg2010recognition,crenn2016body}. Affective features are comprised of two types of features:
\begin{itemize}
    \item Posture features. These include angle and distance between the joints, area of different parts of the body (\textit{e.g.}, area of the triangle formed by the neck, the right hand and the left hand), and the bounding volume of the body.
    \item Movement features. These include the velocity and acceleration of individual joints in the gait.
\end{itemize}
We exploit the affective feature formulation~\cite{gait_psych4,affective} in our final network. We append the $29$ dimensional affective feature~(purple box) to the final layer feature vector learned by our Baseline-STEP network, thus generating hybrid feature vectors. These hybrid feature vectors are passed through two fully connected layers of dimensions $128$ and $4$ respectively, followed by a softmax operation to generate the final class labels. We call this combined network STEP.

\section{Experiments and Results}\label{sec:results}

We list all the parameters and hardware used in training both our generation and classification networks in Section~\ref{subsec:hyperparams}. In Section~\ref{subsec:ewalk}, we give details of our new dataset. In Sections~\ref{subsec:metric}, we list the standard metrics used to compare generative models and classification networks and in Section~\ref{subsec:method}, we list the state-of-the-art methods against which we compare our algorithms. In Section \ref{subsec:results}, we present the evaluation results. Finally, in Section~\ref{overfitting}, we analyse the robustness of our system and show that both STEP and STEP-Gen do not overfit on the E-Gait Dataset.

\subsection{Training Parameters}~\label{subsec:hyperparams}
For training STEP-Gen, we use a batch size of $8$ and train for $150$ epochs. We use the Adam optimizer~\cite{adam} with an initial learning rate of $0.1$, which decreases to $\frac{1}{10}$-th of its current value after $75$, $113$ and $132$ epochs. We also use a momentum of $0.9$ and and weight-decay of $5\times 10^{-4}$.

For training STEP, we use a split of $7:2:1$ for training, validation and testing sets. We use a batch size of $8$ and train for $500$ epochs using the Adam optimizer~\cite{adam} with an initial learning rate of $0.1$. The learning rate decreases to $\frac{1}{10}$-th of its current value after $250$, $375$ and $438$ epochs. We also use a momentum of $0.9$ and and weight-decay of $5\times 10^{-4}$. 
All our results were generated on an NVIDIA GeForce GTX 1080 Ti GPU.

\subsection{Dataset: Emotion-Gait}
\label{subsec:ewalk}
Emotion-Gait (E-Gait) consists of $2,177$ real gaits and $1,000$ synthetic gaits each of the 4 emotion classes generated by STEP-Gen, for a total for $3,177$ gaits. We collected $342$ of the real gaits ourselves. We asked $90$ participants to walk while thinking of the four different emotions (angry, neutral, happy and sad). The total distance of walking for each participant was $7$ meters. The videos were labeled by domain experts. The remaining $1,835$ gaits are taken as is from the Edinburgh Locomotion MOCAP Database~\cite{ELMD}. However, since these gaits did not have any associated labels, we got them labeled with the 4 emotions by the same domain experts.


\begin{figure}[t]
    \centering
    \includegraphics[width=\columnwidth]{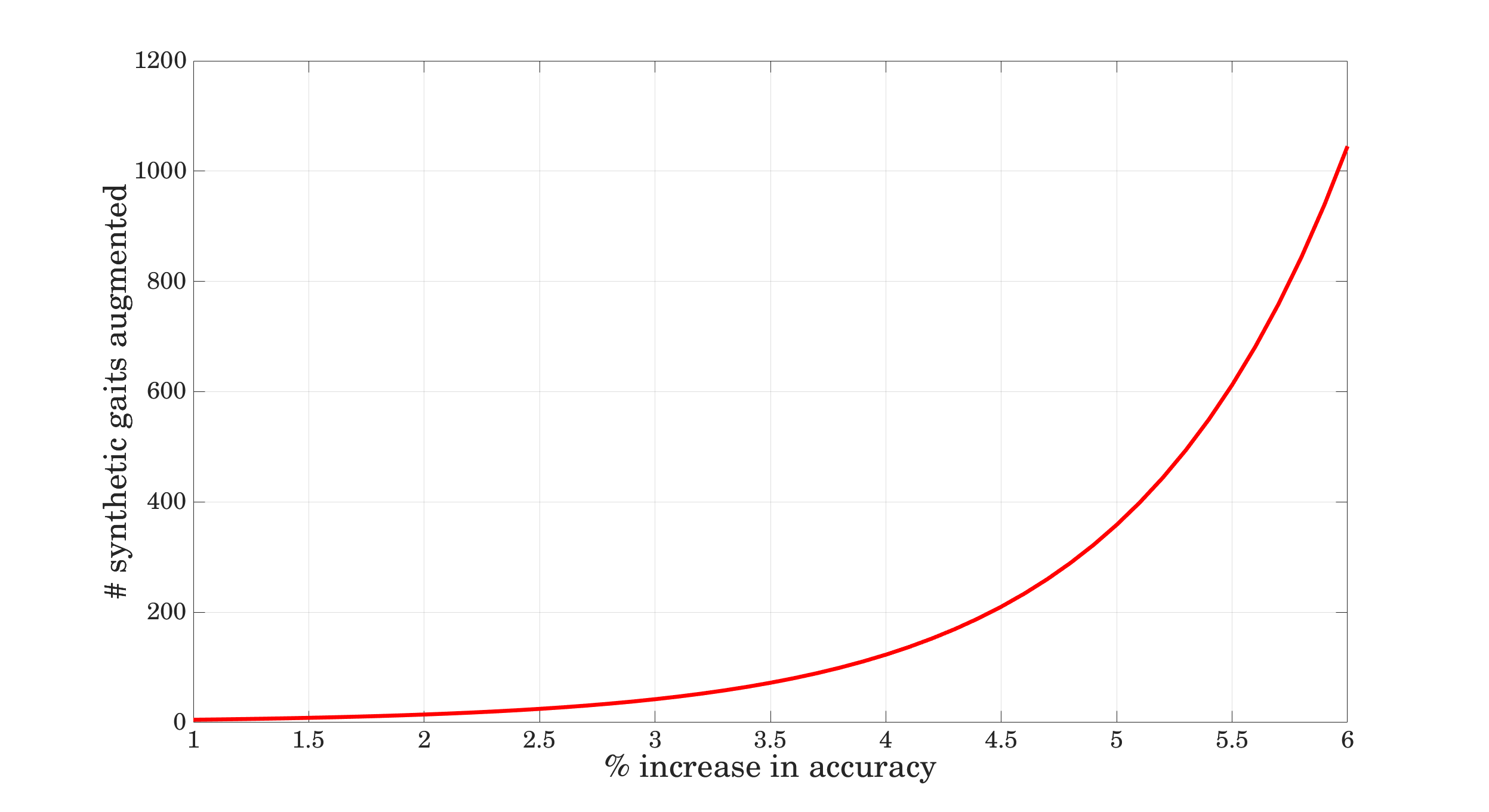}
    \caption{\small\textbf{Effect of Data Augmentation:} Effect of augmenting synthetically generated data to the train and test sets of STEP+Aug on its performance. For every percent improvement in accuracy, an exponentially larger number of data need to be augmented.}
    \label{fig:acc_vs_data}
\end{figure}

\subsection{Evaluation Metrics}
\label{subsec:metric}
\noindent \textbf{Generation:} For generative models, we compute the Fr\'echet Inception Distance (FID) score~\cite{fid} that measures how close the generated samples are to the real inputs while maintaining diversity among the generated samples. The FID score is computed using the following formula:
\begin{equation}
   FID = \lVert \mu_x - \mu_g\rVert_2^2 + Tr(\Sigma_x + \Sigma_g + 2(\Sigma_x\Sigma_g)^{\frac{1}{2}})
\end{equation}

\noindent \textbf{Classification:} For classifier models, we report the classification accuracy given by $Accuracy = (TP + TN)/TD$, where $TP, TN, TD$ are the number of true positives, true negatives, and total data, respectively.

\subsection{Evaluation Methods}
\label{subsec:method}

\noindent \textbf{Generation:} We compare our generative network with both GAN- and VAE-based generative networks, as listed below.

\begin{itemize}
    \item vid2vid (GAN-based)~\cite{vid2vid}: This is the state-of-the-art video generation method. It can take human motion videos as input and generate high-resolution videos of the same motion.
    \item Baseline CVAE (VAE-based): We use a CVAE with the same network architecture as STEP-Gen, but with only the original CVAE loss given in Equation~\ref{equation:originalCVAE} as the reconstruction loss.
\end{itemize}

\noindent \textbf{Classification:} We compare our classifier network with both prior methods for emotion recognition from gaits, and prior methods for action recognition from gaits, as listed below.
\begin{itemize}
    \item Emotion Recognition: We compare with the current state-of-the-art classifiers of~\cite{karg2010recognition,venture2014recognizing,crenn2016body,wang2016adaptive,daoudi2017emotion}.
    \item Action Recognition: We compare with the state-of-the-art methods using both GCNs~\cite{stgcn} and LSTMs~\cite{tanmay_lstm}. The networks of both these methods were trained on our dataset before comparing the performance.
\end{itemize}

We also perform the following ablation experiments with our classifier network:
\begin{itemize}
    \item Baseline-STEP: It predicts emotions based only on the network-learned features from gaits. This network is trained on the $2,177$ real gaits in E-Gait.

    \item STEP: This is our hybrid network combining affective features~\cite{gait_psych4,affective} with the network-learned features of Baseline-STEP. This network is also trained on the $2,177$ real gaits in E-Gait.
    
    \item STEP+Aug: This is the same implementation as STEP, but trained on both the real and the synthetic gaits in E-Gait.
\end{itemize}

\subsection{Results on E-Gait}
\label{subsec:results}
\noindent \textbf{Generation: } All the generative networks are trained on the $2,177$ real data in E-Gait. We report an FID score of $11.11$, while the FID score of Baseline-CVAE is $11.33$. Lower FID indicates higher fidelity to the real data. However, we also note that vid2vid~\cite{vid2vid} completely memorizes the dataset and thus gives an FID score of $0$. This is undesirable for our task since we require the generative network to be able to produce diverse data that can be augmented to the training set of the classifier network.
\begin{figure}[t]
    \centering
    \includegraphics[width=\columnwidth]{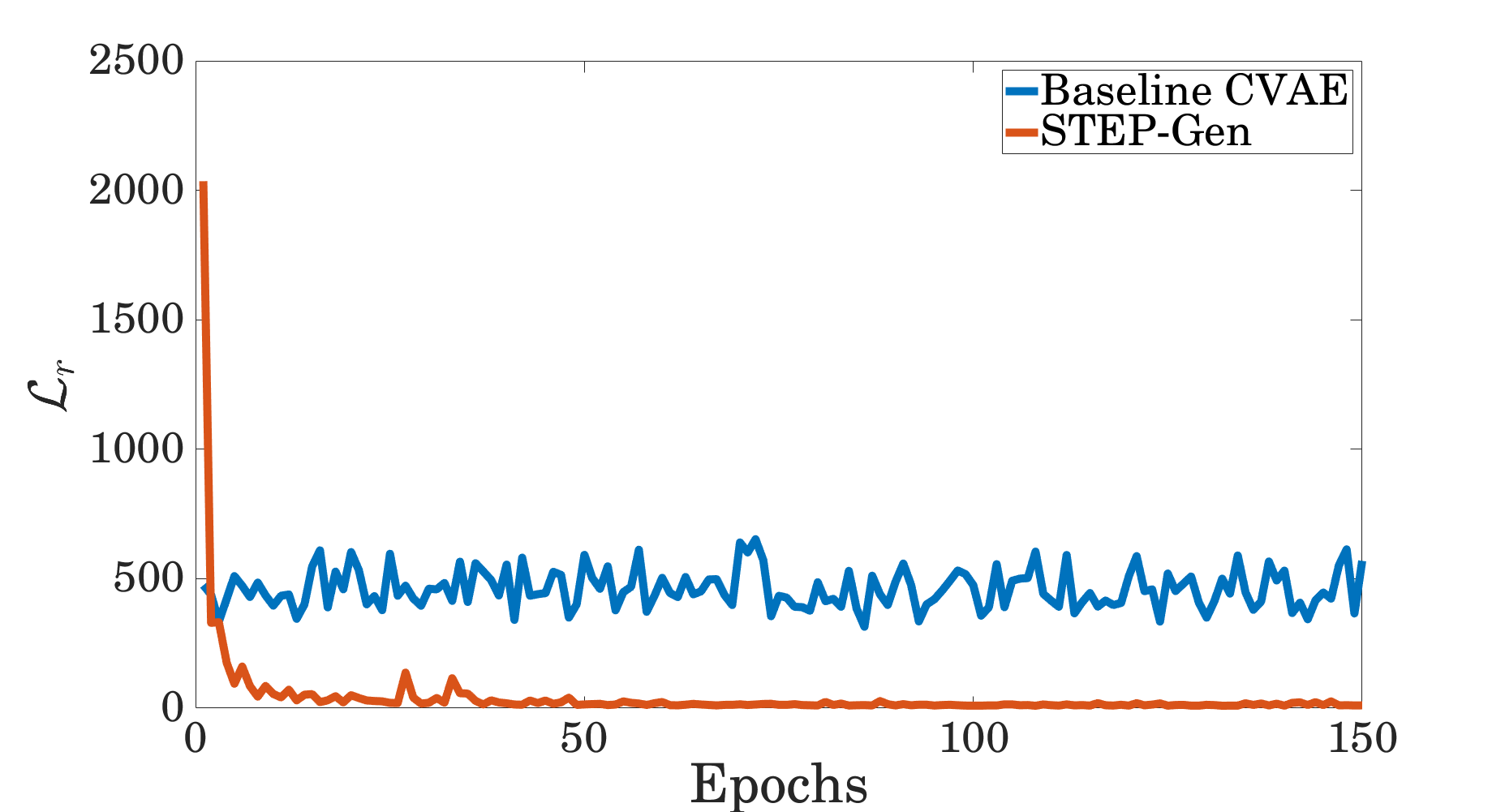}
    \caption{\small\textbf{Training Loss Convergence:} Our ``Push-Pull'' regularization loss (Equation~\ref{equation:STEP-Gen Loss}) as a function of training epochs, as produced by the baseline-CVAE and our STEP-Gen. The baseline-CVAE fails to converge even after $150$ epochs, while STEP-Gen converges after approximately 28 epochs.}
    \label{fig:VAEloss_VS_OurLoss}
\end{figure}


Additionally, to show that our novel ``Push-Pull'' regularization loss function (Equation~\ref{equation:STEP-Gen Loss}) generates gaits with joint movements, we measure the decay of the value of the loss function for the baseline-CVAE and STEP-Gen with time (Figure~\ref{fig:VAEloss_VS_OurLoss}). We add the $\mathcal{L}_c$ and $\mathcal{L}_d$ terms from equation~\ref{equation:STEP-Gen Loss} (without optimizing them) to the baseline-CVAE loss function (Equation~\ref{equation:originalCVAE}). We observe that STEP-Gen converges extremely quickly to a smaller loss value in around 28 epochs. On the other hand, the base-line CVAE produces oscillations and fails to converge as it does not optimize $\mathcal{L}_c$ and $\mathcal{L}_d$.

We also perform qualitative tests of gait generated by all the methods. vid2vid~\cite{vid2vid} uses GANs to produce high-quality videos. However, in our experiments, vid2vid memorizes the dataset and does not produce diverse samples. Baseline-CVAE produces static gaits that do not move in time. Finally, our gaits are both diverse (different from input) and realistic (successfully mimics walking motion). We show all these results in our demo video\footnote{demo video available at: \url{https://gamma.umd.edu/researchdirections/affectivecomputing/step}}.

\begin{table}[!htb]
    \centering
    \resizebox{\columnwidth}{!}{%
    \begin{tabular}{|C{0.9cm}|C{0.9cm}|C{0.9cm}|C{0.9cm}|C{0.9cm}|C{0.9cm}|C{0.9cm}|| C{0.9cm}|C{0.9cm}|C{0.9cm}|}
    \hline
    Venture et al. \cite{venture2014recognizing} & Karg et al. \cite{karg2010recognition} & Daoudi et al. \cite{daoudi2017emotion} & Wang et al. \cite{wang2016adaptive} & Crenn et al. \cite{crenn2016body} & ST-GCN \cite{stgcn} & LSTM \cite{tanmay_lstm} & Base-STEP & STEP & \textbf{STEP + Aug} \\
    \hline
    30.83 & 39.58 & 42.52 & 53.73 & 66.22 & 65.62 & 75.10 & 78.24 & 83.15 & \textbf{89.41} \\
    \hline
  \end{tabular}
  }
  \caption{\textbf{Classification Accuracy Comparison:} Accuracies are computed using the formula in Section~\ref{subsec:metric} and shown in increasing order. We choose methods from both psychology and computer vision literature. Base-STEP and STEP+Aug are variations of STEP.}
  \vspace{-10pt}
  \label{tab:ablations}
\end{table}

\noindent \textbf{Classification:} In Table~\ref{tab:ablations}, we report the mean classification accuracies of all the methods using the formula in Section~\ref{subsec:metric}. We observe that most of the prior methods for emotion recognition from gaits have less than $60\%$ accuracy on E-Gait. Only Crenn et al.~\cite{crenn2016body}, where the authors manually compute the same features we use in our novel ``push-pull'' regularization loss function (enforce \textit{i.e.} distances between joints across time) has greater than $65\%$ accuracy. The two prior action recognition from gait methods we compare with have $65\%$ and $75\%$ accuracy respectively. By comparison, our Baseline-STEP has an accuracy of $78\%$. Combining network-learned and affective features in STEP gives an accuracy of $83\%$. Finally, augmenting synthetic gaits generated by STEP-Gen in STEP+Aug gives an accuracy of $89\%$.

\begin{figure}[t]
    \centering
    \includegraphics[width=\columnwidth]{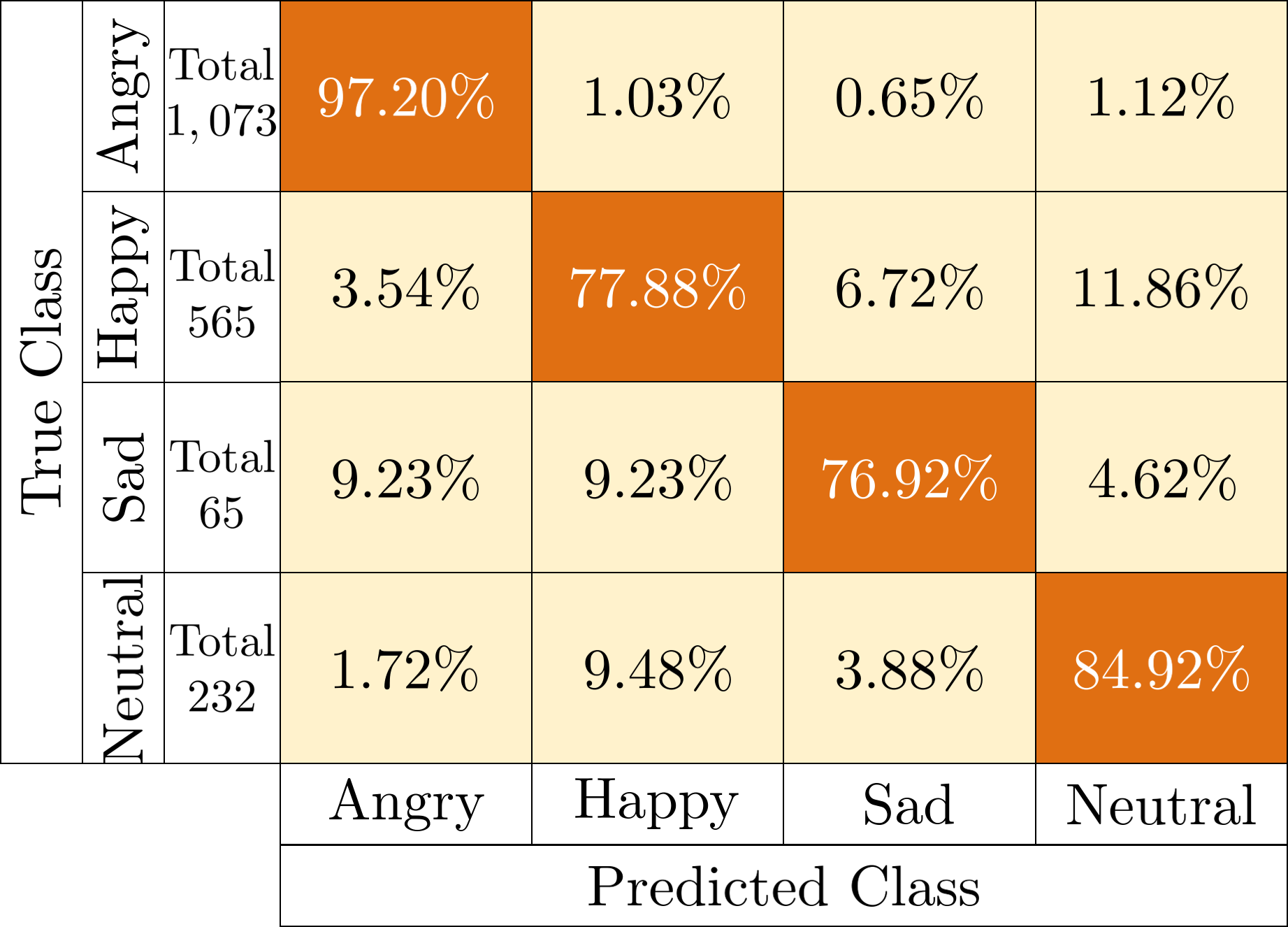}
    \caption{\small\textbf{Accuracy Analysis of STEP+Aug on E-Gait dataset: } Classification results over the $3,177$ gaits. We observe $>75\%$ accuracy for each class.}
    \label{fig:confusion}
\end{figure}


To verify that our classification accuracy is statistically significant and not due to random chance, we perform two statistical tests:

\begin{itemize}
    \item \textit{Hypotheses Testing:} Classification as a task, depends largely on the test sample to be classified. To ensure that the classification accuracy of STEP is not achieved due to random positive examples, we determine the statistical likelihood of our results. Note that we do not test on STEP+Aug as accuracy of STEP+Aug is also dependent on the augmentation size. We generate a population of size $10,000$ accuracy values of STEP with mean $83.15$ and standard deviation $6.9$. We set $\mu = 83.15$, \textit{i.e.} the reported mean accuracy of STEP as the null hypothesis, $H_0$. To accept our null hypothesis, we require the p-value to be greater than $0.50$. We compute the p-value of this population as $0.78>0.50$. Therefore, we fail to reject the null hypothesis, thus corroborating our classification accuracy statistically.
    
    \item \textit{Confidence Intervals:} This metric determines the likelihood of a value residing in an interval. For a result to be meaningful and statistically significant, we require a tight interval with high probability. With a $95\%$ likelihood, we report a confidence interval of $[81.19, 85.33]$ with a standard deviation of $1.96$. Simply put, our classification accuracy will lie between $81.19$ and $85.33$ with a probability of $0.95$.
\end{itemize}

Finally, we show the discriminatory capability of our classifier through a confusion matrix in Figure~\ref{fig:confusion}.

\subsection{Overfitting Analysis}\label{overfitting}
\noindent \textbf{Effect of Generated Data on Classification: } We show in Figure~\ref{fig:acc_vs_data} that the synthetic data generated by STEP-Gen increases the classification accuracy of STEP+Aug. This, in turn, shows that STEP-Gen does not memorize the training dataset, but can produce useful diverse samples. Nevertheless, we see that to achieve every percent improvement in the accuracy of STEP+Aug, we need to generate an exponentially larger number of synthetic samples as training saturation sets in.

\begin{figure}[t]
    \centering
    \includegraphics[width=\columnwidth]{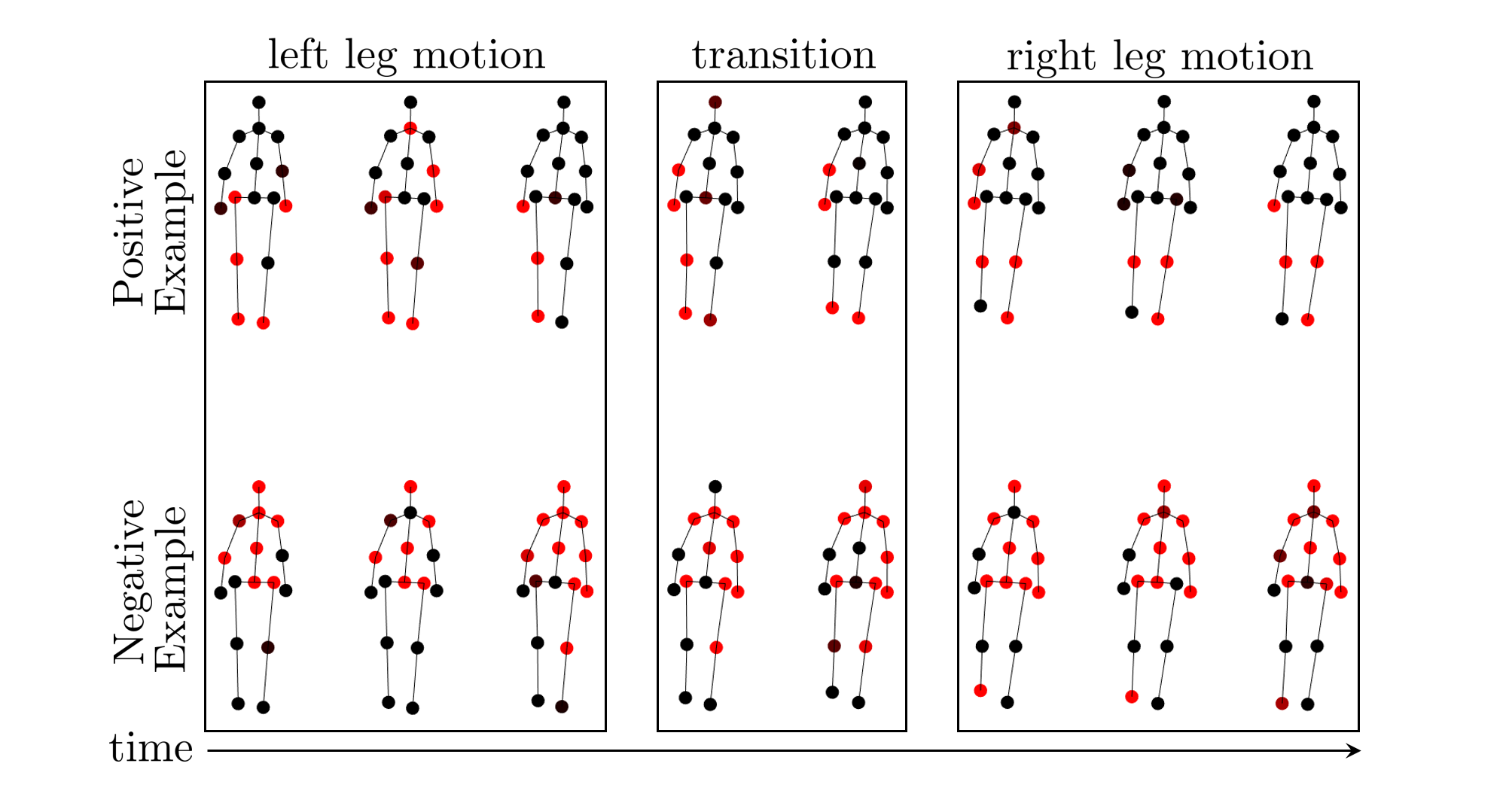}
    \caption{\small\textbf{Saliency Map:} Saliency map showing the magnitude of the network gradient along the $z$-axis (in and out of the paper) generated by our trained network, which is the direction of walking in both the examples shown. The examples are for the `happy' emotion. In the positive example, the network correctly detects simultaneous movement in the right leg and the left hand, followed by a transition period, followed by simultaneous movement of the left leg and the right hand. These movements are important emotional cues, thereby confirming that our classifier is learning meaningful features to recognize emotions accurately.}
    \label{fig:saliency}
\end{figure}

\noindent \textbf{Saliency Maps: }We show that STEP does not memorize the training dataset, but learns meaningful features, using saliency maps obtained via guided backpropagation on the learned network~\cite{saliency1,saliency2}. Saliency maps determine how the loss function output changes with respect to a small change in the input. In our case, the input consists of 3D joint positions over time, therefore, the corresponding saliency map highlights the joints that cause the most influence the output. Intuitively, we expect the saliency map for a positively classified example to capture the joint movements that are most important for predicting the perceived emotion from a psychological point of view~\cite{crenn2016body}. We show the saliency map given by our trained network for both a positively classified and a negatively classified example for the label `happy' in Figure~\ref{fig:saliency}. The saliency map only shows magnitude of the gradient along the $z$-axis (in and out of the plane of the paper), which is the direction of walking in both the examples. Black represents zero magnitude, and bright red represents a high magnitude. In the positive example, we see that the network detects simultaneous movement in the right leg and the left hand, followed by a transition period, followed by simultaneous movement of the left leg and the right hand. This is the expected behavior, as the movement of hands and the stride length and speed are important cues for emotion perception~\cite{crenn2016body}. Note that other movements, such as that of the spine, lie along the other axes directions, and hence are not captured in the shown saliency map. By the contrast, there is no intuitive pattern to the detected movements in the saliency map for the negative example. For the sake of completeness, we provide the saliency maps along the other axes directions in the demo video.\\

\section{Limitations and Future Work}
Our generative model is currently limited to generating gait sequences of a single person. The accuracy of the classification algorithm is also governed by the quality of the video and the pose extraction algorithm. 
There are many avenues for future work as well. We would like to extend the approach to deal with multi-person or crowd videos. Given the complexity of generating annotated real-world videos, we need better generators to improve the accuracy of classification algorithm. Lastly, it would be useful to combine gait-based emotion classification with other modalities corresponding to face-expressions or speech to further improve the accuracy.

\bibliographystyle{abbrv}
\bibliography{step_refs}

\begin{thebibliography}{10}

\bibitem{crossmodal2}
S.~Albanie, A.~Nagrani, A.~Vedaldi, and A.~Zisserman.
\newblock Emotion recognition in speech using cross-modal transfer in the wild.
\newblock {\em arXiv:1808.05561}, 2018.

\bibitem{surveillance2}
J.~Arunnehru and M.~K. Geetha.
\newblock Automatic human emotion recognition in surveillance video.
\newblock In {\em ITSPMS}, pages 321--342. Springer, 2017.

\bibitem{affective2}
M.~Atcheson, V.~Sethu, and J.~Epps.
\newblock Gaussian process regression for continuous emotion recognition with
  global temporal invariance.
\newblock In {\em IJCAI-W}, pages 34--44, 2017.

\bibitem{bauer2009autonomous}
A.~Bauer et~al.
\newblock The autonomous city explorer: Towards natural human-robot interaction
  in urban environments.
\newblock {\em IJSR}, 1(2):127--140, 2009.

\bibitem{gcn1}
J.~Bruna, W.~Zaremba, A.~Szlam, and Y.~LeCun.
\newblock Spectral networks and locally connected networks on graphs.
\newblock {\em arXiv:1312.6203}, 2013.

\bibitem{pose_guided_gait2}
H.~Cai, C.~Bai, Y.-W. Tai, and C.-K. Tang.
\newblock Deep video generation, prediction and completion of human action
  sequences.
\newblock In {\em ECCV}, pages 366--382, 2018.

\bibitem{gait_emo}
M.~Chiu, J.~Shu, and P.~Hui.
\newblock Emotion recognition through gait on mobile devices.
\newblock In {\em PerCom Workshops}, pages 800--805. IEEE, 2018.

\bibitem{crenn2016body}
A.~Crenn, R.~A. Khan, A.~Meyer, and S.~Bouakaz.
\newblock Body expression recognition from animated 3d skeleton.
\newblock In {\em IC3D}, pages 1--7. IEEE, 2016.

\bibitem{affective}
A.~Crenn, R.~A. Khan, A.~Meyer, and S.~Bouakaz.
\newblock Body expression recognition from animated 3d skeleton.
\newblock In {\em IC3D}, pages 1--7. IEEE, 2016.

\bibitem{pose_extract_1}
R.~Dabral, A.~Mundhada, U.~Kusupati, S.~Afaque, A.~Sharma, and A.~Jain.
\newblock Learning 3d human pose from structure and motion.
\newblock {\em Computer Vision -- ECCV 2018}, 2018.

\bibitem{daoudi2017emotion}
M.~Daoudi, S.~Berretti, P.~Pala, Y.~Delevoye, and A.~Del~Bimbo.
\newblock Emotion recognition by body movement representation on the manifold
  of symmetric positive definite matrices.
\newblock In {\em ICIAP}, pages 550--560. Springer, 2017.

\bibitem{speech_deep3}
J.~Deng, X.~Xu, Z.~Zhang, S.~Fr{\"u}hholz, and B.~Schuller.
\newblock Semisupervised autoencoders for speech emotion recognition.
\newblock {\em IEEE/ACM ASLP}, 26(1):31--43, 2018.

\bibitem{behavior2}
S.~A. Denham, E.~Workman, P.~M. Cole, C.~Weissbrod, K.~T. Kendziora, and
  C.~ZAHN-WAXLER.
\newblock Prediction of externalizing behavior problems from early to middle
  childhood.
\newblock {\em Development and Psychopathology}, 12(1):23--45, 2000.

\bibitem{face_re}
P.~Ekman.
\newblock Facial expression and emotion.
\newblock {\em American psychologist}, 48(4):384, 1993.

\bibitem{ekman1967head}
P.~Ekman and W.~V. Friesen.
\newblock Head and body cues in the judgment of emotion: A reformulation.
\newblock {\em Perceptual and motor skills}, 1967.

\bibitem{AU}
C.~Fabian Benitez-Quiroz, R.~Srinivasan, and A.~M. Martinez.
\newblock Emotionet: An accurate, real-time algorithm for the automatic
  annotation of a million facial expressions in the wild.
\newblock In {\em CVPR}, June 2016.

\bibitem{facial_deep2}
Y.~Fan, X.~Lu, D.~Li, and Y.~Liu.
\newblock Video-based emotion recognition using cnn-rnn and c3d hybrid
  networks.
\newblock In {\em ICMI}, pages 445--450. ACM, 2016.

\bibitem{i3c}
C.~Feichtenhofer, A.~Pinz, and R.~Wildes.
\newblock Spatiotemporal residual networks for video action recognition.
\newblock In {\em NIPS}, pages 3468--3476, 2016.

\bibitem{twostream}
C.~Feichtenhofer, A.~Pinz, and A.~Zisserman.
\newblock Convolutional two-stream network fusion for video action recognition.
\newblock In {\em CVPR}, pages 1933--1941, 2016.

\bibitem{face_audience}
J.-M. Fern{\'a}ndez-Dols and M.-A. Ruiz-Belda.
\newblock Expression of emotion versus expressions of emotions.
\newblock In {\em Everyday conceptions of emotion}, pages 505--522. Springer,
  1995.

\bibitem{pose_extract_2}
R.~Girdhar, G.~Gkioxari, L.~Torresani, M.~Paluri, and D.~Tran.
\newblock Detect-and-track: Efficient pose estimation in videos.
\newblock {\em CoRR}, abs/1712.09184, 2017.

\bibitem{ELMD}
I.~Habibie, D.~Holden, J.~Schwarz, J.~Yearsley, and T.~Komura.
\newblock A recurrent variational autoencoder for human motion synthesis.
\newblock In {\em {BMVC}}, 2017.

\bibitem{fid}
M.~Heusel, H.~Ramsauer, T.~Unterthiner, B.~Nessler, and S.~Hochreiter.
\newblock Gans trained by a two time-scale update rule converge to a local nash
  equilibrium.
\newblock In {\em NIPS}, pages 6626--6637, 2017.

\bibitem{prosodic2}
A.~Jacob and P.~Mythili.
\newblock Prosodic feature based speech emotion recognition at segmental and
  supra segmental levels.
\newblock In {\em SPICES}, pages 1--5. IEEE, 2015.

\bibitem{c3d}
S.~Ji, W.~Xu, M.~Yang, and K.~Yu.
\newblock 3d convolutional neural networks for human action recognition.
\newblock {\em PAMI}, 35(1):221--231, 2013.

\bibitem{karg2010recognition}
M.~Karg, K.~Kuhnlenz, and M.~Buss.
\newblock Recognition of affect based on gait patterns.
\newblock {\em Cybernetics}, 40(4):1050--1061, 2010.

\bibitem{adam}
D.~P. Kingma and J.~Ba.
\newblock Adam: A method for stochastic optimization.
\newblock {\em arXiv:1412.6980}, 2014.

\bibitem{vae}
D.~P. Kingma and M.~Welling.
\newblock Auto-encoding variational bayes.
\newblock {\em arXiv:1312.6114}, 2013.

\bibitem{gcn2}
T.~N. Kipf and M.~Welling.
\newblock Semi-supervised classification with graph convolutional networks.
\newblock {\em arXiv:1609.02907}, 2016.

\bibitem{gait_psych4}
A.~Kleinsmith and N.~Bianchi-Berthouze.
\newblock Affective body expression perception and recognition: A survey.
\newblock {\em IEEE Transactions on Affective Computing}, 4(1):15--33, 2013.

\bibitem{cnn_alexnet}
A.~Krizhevsky, I.~Sutskever, and G.~E. Hinton.
\newblock Imagenet classification with deep convolutional neural networks.
\newblock In {\em NIPS}, pages 1097--1105, 2012.

\bibitem{tcn}
C.~Lea, M.~D. Flynn, R.~Vidal, A.~Reiter, and G.~D. Hager.
\newblock Temporal convolutional networks for action segmentation and
  detection.
\newblock In {\em CVPR}, pages 156--165, 2017.

\bibitem{facial}
A.~Majumder, L.~Behera, and V.~K. Subramanian.
\newblock Emotion recognition from geometric facial features using
  self-organizing map.
\newblock {\em Pattern Recognition}, 47(3):1282--1293, 2014.

\bibitem{gait_psych2}
H.~K. Meeren, C.~C. van Heijnsbergen, and B.~de~Gelder.
\newblock Rapid perceptual integration of facial expression and emotional body
  language.
\newblock {\em Proceedings of NAS}, 102(45):16518--16523, 2005.

\bibitem{gait_psych3}
J.~Michalak, N.~F. Troje, J.~Fischer, P.~Vollmar, T.~Heidenreich, and
  D.~Schulte.
\newblock Embodiment of sadness and depression—gait patterns associated with
  dysphoric mood.
\newblock {\em Psychosomatic Medicine}, 71(5):580--587, 2009.

\bibitem{gait_psych1}
J.~M. Montepare, S.~B. Goldstein, and A.~Clausen.
\newblock The identification of emotions from gait information.
\newblock {\em Journal of Nonverbal Behavior}, 11(1):33--42, 1987.

\bibitem{perception_imp}
K.~S. Quigley, K.~A. Lindquist, and L.~F. Barrett.
\newblock Inducing and measuring emotion and affect: Tips, tricks, and secrets.
\newblock {\em Cambridge University Press}, 2014.

\bibitem{tanmay_lstm}
T.~Randhavane, A.~Bera, K.~Kapsaskis, U.~Bhattacharya, K.~Gray, and D.~Manocha.
\newblock Identifying emotions from walking using affective and deep features.
\newblock {\em arXiv:1906.11884}, 2019.

\bibitem{eye2}
M.~Schurgin, J.~Nelson, S.~Iida, H.~Ohira, J.~Chiao, and S.~Franconeri.
\newblock Eye movements during emotion recognition in faces.
\newblock {\em Journal of vision}, 14(13):14--14, 2014.

\bibitem{saliency1}
K.~Simonyan and A.~Zisserman.
\newblock Very deep convolutional networks for large-scale image recognition.
\newblock {\em arXiv:1409.1556}, 2014.

\bibitem{cvae}
K.~Sohn, H.~Lee, and X.~Yan.
\newblock Learning structured output representation using deep conditional
  generative models.
\newblock In {\em NIPS}, pages 3483--3491, 2015.

\bibitem{saliency2}
J.~T. Springenberg, A.~Dosovitskiy, T.~Brox, and M.~Riedmiller.
\newblock Striving for simplicity: The all convolutional net.
\newblock {\em arXiv:1412.6806}, 2014.

\bibitem{mocogan}
S.~Tulyakov, M.-Y. Liu, X.~Yang, and J.~Kautz.
\newblock Mocogan: Decomposing motion and content for video generation.
\newblock In {\em CVPR}, pages 1526--1535, 2018.

\bibitem{umeyama}
S.~Umeyama.
\newblock Least-squares estimation of transformation parameters between two
  point patterns.
\newblock {\em TPAMI}, pages 376--380, 1991.

\bibitem{venture2014recognizing}
G.~Venture, H.~Kadone, T.~Zhang, J.~Gr{\`e}zes, A.~Berthoz, and H.~Hicheur.
\newblock Recognizing emotions conveyed by human gait.
\newblock {\em IJSR}, 6(4):621--632, 2014.

\bibitem{acrec2}
L.~Wang, T.~Tan, W.~Hu, and H.~Ning.
\newblock Automatic gait recognition based on statistical shape analysis.
\newblock {\em TIP}, 12(9):1120--1131, 2003.

\bibitem{vid2vid}
T.-C. Wang, M.-Y. Liu, J.-Y. Zhu, G.~Liu, A.~Tao, J.~Kautz, and B.~Catanzaro.
\newblock Video-to-video synthesis.
\newblock In {\em NeurIPS}, 2018.

\bibitem{wang2016adaptive}
W.~Wang, V.~Enescu, and H.~Sahli.
\newblock Adaptive real-time emotion recognition from body movements.
\newblock {\em TiiS}, 5(4):18, 2016.

\bibitem{stgcn}
S.~Yan, Y.~Xiong, and D.~Lin.
\newblock Spatial temporal graph convolutional networks for skeleton-based
  action recognition.
\newblock In {\em AAAI}, 2018.

\bibitem{pose_guided_gait1}
C.~Yang, Z.~Wang, X.~Zhu, C.~Huang, J.~Shi, and D.~Lin.
\newblock Pose guided human video generation.
\newblock In {\em ECCV}, pages 201--216, 2018.

\bibitem{facial_deep3}
H.~Yang, U.~Ciftci, and L.~Yin.
\newblock Facial expression recognition by de-expression residue learning.
\newblock In {\em CVPR}, pages 2168--2177, 2018.

\bibitem{affective1}
H.~Yates, B.~Chamberlain, G.~Norman, and W.~H. Hsu.
\newblock Arousal detection for biometric data in built environments using
  machine learning.
\newblock In {\em IJCAI-W}, pages 58--72, 2017.

\bibitem{facial_deep4}
F.~Zhang, T.~Zhang, Q.~Mao, and C.~Xu.
\newblock Joint pose and expression modeling for facial expression recognition.
\newblock In {\em CVPR}, pages 3359--3368, 2018.

\bibitem{identification2}
Z.~Zhang and N.~F. Troje.
\newblock View-independent person identification from human gait.
\newblock {\em Neurocomputing}, 69(1-3):250--256, 2005.

\bibitem{physiological2}
M.~Zhao, F.~Adib, and D.~Katabi.
\newblock Emotion recognition using wireless signals.
\newblock In {\em ICMCN}, pages 95--108. ACM, 2016.

\end{thebibliography}
\end{document}